\documentclass{decar-wsd}    

\usepackage{decar-common}    
\usepackage{decar-dynamics}  
\usepackage{decar-lie}       
\usepackage{decar-post}      

\usepackage{subfig}
\usepackage{graphicx}
\usepackage{stfloats}

\usepackage{amsmath,amsfonts}
\usepackage{algorithmic}
\usepackage{algorithm}
\usepackage{array}
\usepackage{textcomp}
\usepackage{url}
\usepackage{verbatim}
\usepackage{graphicx}
\usepackage{cite}
\usepackage{adjustbox}
\usepackage{lipsum}  

\usepackage[most]{tcolorbox}

\makeatletter
\newcommand{\vast}{\bBigg@{4}}
\newcommand{\Vast}{\bBigg@{5}}
\makeatother

\newcounter{comment}[section]

\title{IMU Preintegration for Multi-Robot Systems in the Presence of Bias and Communication Constraints}
\author{Mohammed Ayman Shalaby, Charles Champagne Cossette, Jerome Le Ny, James Richard Forbes}

\begin{document}

\maketitle

\section*{Abstract}

This document is in supplement to the paper titled ``Multi-Robot Relative Pose Estimation and IMU Preintegration Using Passive UWB Transceivers'', available at \cite{Shalaby2023c}. The purpose of this document is to show how IMU biases can be incorporated into the framework presented in \cite{Shalaby2023c}, while maintaining the differential Sylvester equation form of the process model.

\section{Introduction}

The need to estimate IMU biases is particularly important for applications that involve long-term navigation. As compared to the framework presented in \cite{Shalaby2023c}, most robotic applications involve additional exteroceptive sensors such as GPS, a magnetometer, or a camera that provide measurements relative to static environmental quantities, which allow individual robots to estimate their own IMU biases using standard methods \cite[Ch. 10]{Farrell2008}, \cite{Hong2005,Mourikis2007}. In a multi-robot scenario, these bias estimates can be used by each robot to correct its own IMU measurement before adding the measurement to the RMI. This is a loosely-coupled solution that overcomes the need for each robot to share its IMU biases with its neighbours. 

In \cite{Shalaby2023c}, only range measurements are available that provide constraints among two moving bodies, hence estimating the biases is trickier as there is no static reference. Nonetheless, even though the proposed framework will mostly be used alongside additional exteroceptive sensors such as a camera to allow for real-world applications, it is indeed important to address the issue of IMU biases in the context of the framework presented in \cite{Shalaby2023c} to allow long-term navigation without relying on additional sensors. To do so, this document presents how gyroscope biases and relative accelerometer biases can be estimated while maintaining the differential Sylvester equation form of the process model, under a few assumptions. 

The remainder of this document uses the same notation as in \cite{Shalaby2023c}, and is organized as follows. Section \ref{sec:pose_process} presents the pose process model with IMU biases, and Section \ref{sec:bias_process} presents the bias process model. In Section \ref{sec:preintegration}, preintegration of the IMU measurements. The simulation and experimental results are shown in Sections \ref{sec:simulation} and \ref{sec:experimental}, respectively.

\section{Pose Process Model with Bias} \label{sec:pose_process}

The IMU biases affect the process model presented in \cite{Shalaby2023c}, but the ranging protocol and the form of the measurement model remain unchanged. The relative attitude process model is 
\begin{align}
    \mbfdot{C}_{0i} &= - \left( \mbs{\omega}_0^{0a} - \mbs{\beta}_0^\text{gyr,0} \right)^\times \mbf{C}_{0i} + \mbf{C}_{0i} \left( \mbs{\omega}_i^{ia} - \mbs{\beta}_i^\text{gyr,i} \right)^\times
\end{align}
in the presence of bias, where $\mbs{\beta}_i^\text{gyr,i}$ is the bias of the gyroscope of Robot $i$ resolved in the Robot $i$'s body frame. Similarly, the relative velocity process model is
\begin{align}
    ^0 \mbfdot{v}_0^{i0/a} &= - \left(\mbs{\omega}_0^{0a} - \mbs{\beta}_0^\text{gyr,0} \right)^\times \mbf{v}_0^{i0/a} + \mbf{C}_{0i} \left(\mbs{\alpha}_i^i - \mbs{\beta}_i^\text{acc,i}\right) - (\mbs{\alpha}_0^0 - \mbs{\beta}_0^\text{acc,0}) \\
    &= - \left(\mbs{\omega}_0^{0a} - \mbs{\beta}_0^\text{gyr,0} \right)^\times \mbf{v}_0^{i0/a} + \mbf{C}_{0i} \mbs{\alpha}_i^i - \mbs{\alpha}_0^0 \underbrace{- \hspace{2pt} \mbf{C}_{0i} \mbs{\beta}_i^\text{acc,i} + \mbs{\beta}_0^\text{acc,0}}_{\mbs{\beta}_0^{\text{acc},0i}} \\
    &= - \left(\mbs{\omega}_0^{0a} - \mbs{\beta}_0^\text{gyr,0} \right)^\times \mbf{v}_0^{i0/a} + \mbf{C}_{0i} \mbs{\alpha}_i^i - \mbs{\alpha}_0^0 + \mbs{\beta}_0^{\text{acc},0i},
\end{align}
where $\mbs{\beta}_i^\text{acc,i}$ is the bias of the accelerometer of Robot $i$ resolved in the Robot $i$'s body frame, and $\mbs{\beta}_0^\text{acc,0i} \triangleq \mbs{\beta}_0^\text{acc,0} - \mbf{C}_{0i} \mbs{\beta}_i^\text{acc,i}$ is the relative accelerometer bias of Robot 0 relative to Robot $i$, resolved in Robot $0$'s body frame. Lastly, the relative position process model is
\begin{align}
    ^0 \mbfdot{r}_0^{i0} &= - \left(\mbs{\omega}_0^{0a} - \mbs{\beta}_0^\text{gyr,0} \right)^\times \mbf{r}_0^{i0} + \mbf{v}_0^{i0/a}.
\end{align}
Note that the choice of estimating the relative accelerometer bias $\mbs{\beta}_0^\text{acc,0i}$ and the absolute gyroscope bias $\mbs{\beta}_0^\text{gyr,0}$ is made to ensure that the process model remains of the form of a differential Sylvester equation,
\begin{align}
    \mbfdot{T}_{0i} &= \bma{ccc}
        \mbfdot{C}_{0i} & ^0\mbfdot{v}_0^{i0/a} & ^0\mbfdot{r}_0^{i0} \\
        & 0 & \\
        & & 0 
    \ema \nonumber \\
    &= - \left( \bma{ccc}
        \left(\mbs{\omega}_0^{0a}\right)^\times & \mbs{\alpha}_0^0 &  \\
        & & 1 \\
        & & 0 
    \ema - \bma{ccc}
        \left(\mbs{\beta}_0^\text{gyr,0}\right)^\times & \mbs{\beta}_0^{\text{acc},0i} &  \\
        & & 0 \\
        & & 0 
    \ema \right) \mbf{T}_{0i} \nonumber \\ &\hspace{62pt} + \mbf{T}_{0i} \left( \bma{ccc}
        \left(\mbs{\omega}_i^{ia}\right)^\times & \mbs{\alpha}_i^i &  \\
        & & 1 \\
        & & 0 
    \ema - \bma{ccc}
        \left(\mbs{\beta}_i^\text{gyr,i}\right)^\times & \mbf{0} &  \\
        & & 0 \\
        & & 0 
    \ema \right) \nonumber \\
    &\triangleq - (\mbftilde{U}_0 - \mbftilde{B}_0) \mbf{T}_{0i} + \mbf{T}_{0i} (\mbftilde{U}_i - \mbftilde{B}_i). \label{eq:process_model_ct}
\end{align}
This is of a similar form as the process model presented in \cite{Shalaby2023c}, and has a closed-form solution of the form 
\beq
    \label{eq:process_model_dt_wBias}
    \mbf{T}_{0i,k+1} = \underbrace{\operatorname{exp} ((\mbftilde{U}_{0,k} - \mbftilde{B}_{0,k}) \Delta t)^{-1}}_{\mbf{B}_{0,k}^{-1}} \mbf{T}_{0i,k} \underbrace{\operatorname{exp} ((\mbftilde{U}_{i,k} - \mbftilde{B}_{i,k}) \Delta t)}_{\mbf{B}_{i,k}}
\eeq
for an initial condition $\mbf{T}_{0i,k}$. 

The next step is then to linearize the discrete-time process model, in a manner similar to \cite[Section~VI.C]{Shalaby2023c}. Defining $\mbs{\Omega}^{B}_{0,k} \triangleq \left(\mbs{\omega}_{0,k}^{0a} - \mbs{\beta}_{0,k}^\text{gyr,0} \right) \Delta t$ yields
\begin{align*}
    \mbf{B}_{0,k} &= \bma{ccc}
        \operatorname{Exp}(\mbs{\Omega}^{B}_{0,k}) & \Delta t \mbf{J}_l\left(\mbs{\Omega}^\text{B}_{0,k}\right) \left(\mbs{\alpha}_{0,k}^0 - \mbs{\beta}_{0,k}^{\text{acc},0i}\right) & \f{\Delta t^2}{2} \mbf{N}\left(\mbs{\Omega}^\text{B}_{0,k}\right) \left(\mbs{\alpha}_{0,k}^0 - \mbs{\beta}_{0,k}^{\text{acc},0i}\right) \\
        & 1 & \Delta t \\
        & & 1
    \ema \\
    &= \mbf{M} \operatorname{Exp} \vast( \underbrace{\bma{cc}
        \Delta t \mbf{1}  & \\
        & \Delta t \mbf{1} \\
        & \f{\Delta t^2}{2} \mbf{J}_l\left(\mbs{\Omega}^\text{B}_{0,k}\right)^{-1} \mbf{N}\left(\mbs{\Omega}^\text{B}_{0,k}\right)
    \ema}_{\mbf{V}^\text{B}_{0,k}} \Bigg( \underbrace{\bma{c}
        \mbs{\omega}_{0,k}^{0a} \\
        \mbs{\alpha}_{0,k}^0
    \ema}_{\mbf{u}_{0,k}} - \underbrace{\bma{c}
        \mbs{\beta}_{0,k}^\text{gyr,0} \\
        \mbs{\beta}_{0,k}^{\text{acc},0i}
    \ema}_{\mbs{\beta}_{0,k}} \Bigg) \vast) \nonumber \\
    &\triangleq \mbf{M} \operatorname{Exp} \left(\mbf{V}^\text{B}_{0,k} (\mbf{u}_{0,k} - \mbs{\beta}_{0,k})\right).
\end{align*} 
Perturbing this with respect to the input $\mbf{u}_{0,k}$ and the bias $\mbs{\beta}_{0,k}$ yields
\begin{align*}
    \mbf{B}_{0,k} &= \mbfbar{B}_{0,k} \operatorname{Exp} \left(\mbf{L}^\text{B}_{0,k} (\mbfdel{u}_{0,k} - \mbsdel{\beta}_{0,k})\right),
\end{align*}
where $\mbf{L}_{0,k}^\text{B} \triangleq \mbc{J}_l(-\mbfbar{V}^\text{B}_{0,k} (\mbfbar{u}_{0,k} - \mbsbar{\beta}_{0,k}))\mbfbar{V}^\text{B}_{0,k}$, and $\mbc{J}_l(\cdot)$ is the left Jacobian of $SE_2(3)$.
Similarly, defining $\mbs{\Omega}^{B}_{i,k} \triangleq \left(\mbs{\omega}_{i,k}^{ia} - \mbs{\beta}_{i,k}^\text{gyr,i} \right) \Delta t$,  
\begin{align*}
    \mbf{B}_{i,k} &= \bma{ccc}
        \operatorname{Exp}(\mbs{\Omega}^{B}_{i,k}) & \Delta t \mbf{J}_l\left(\mbs{\Omega}^\text{B}_{i,k}\right) \mbs{\alpha}_{i,k}^i & \f{\Delta t^2}{2} \mbf{N}\left(\mbs{\Omega}^\text{B}_{i,k}\right) \mbs{\alpha}_{i,k}^i \\
        & 1 & \Delta t \\
        & & 1
    \ema \\
    &= \mbf{M} \operatorname{Exp} \vast( \underbrace{\bma{cc}
        \Delta t \mbf{1}  & \\
        & \Delta t \mbf{1} \\
        & \f{\Delta t^2}{2} \mbf{J}_l\left(\mbs{\Omega}^\text{B}_{i,k}\right)^{-1} \mbf{N}\left(\mbs{\Omega}^\text{B}_{i,k}\right)
    \ema}_{\mbf{V}^\text{B}_{i,k}} \Bigg( \underbrace{\bma{c}
        \mbs{\omega}_{i,k}^{ia} \\
        \mbs{\alpha}_{i,k}^i
    \ema}_{\mbf{u}_{i,k}} - \underbrace{\bma{c}
        \mbf{1} \\
        \mbf{0}
    \ema}_{\mbf{E}} \mbs{\beta}_{i,k}^\text{gyr,i} \Bigg) \vast) \nonumber \\
    &\triangleq \mbf{M} \operatorname{Exp} \left(\mbf{V}^\text{B}_{i,k} (\mbf{u}_{i,k} - \mbf{E} \mbs{\beta}_{i,k}^\text{gyr,i})\right).
\end{align*} 
Perturbing this with respect to the input $\mbf{u}_{i,k}$ and the bias $\mbs{\beta}_{i,k}^\text{gyr,i}$ yields
\begin{align}
    \label{eq:Bi_perturbation}
    \mbf{B}_{i,k} &= \mbfbar{B}_{i,k} \operatorname{Exp} \left(\mbf{L}^\text{B}_{i,k} (\mbfdel{u}_{i,k} - \mbf{E} \mbsdel{\beta}_{i,k}^\text{gyr,i})\right),
\end{align}
where $\mbf{L}_{i,k}^\text{B} \triangleq \mbc{J}_l(-\mbfbar{V}^\text{B}_{i,k} (\mbfbar{u}_{i,k} - \mbsbar{\beta}_{i,k}^\text{gyr,i}))\mbfbar{V}^\text{B}_{i,k}$.

\section{Bias Process Model} \label{sec:bias_process}

Having derived and linearized the pose process model, the focus now shifts to the bias process model. The bias states being estimated are the gyroscope biases $\mbs{\beta}_0^\text{gyr,0}$ and $\mbs{\beta}_i^\text{gyr,i}$, and the relative accelerometer bias $\mbs{\beta}_0^\text{acc,0i}$. The evolution of IMU biases is oftentimes modelled as a random walk \cite{Forster2017,Mourikis2007}. Therefore, the process model for the gyroscope biases is given by 
\begin{align}
    \mbs{\beta}_{0,k+1}^\text{gyr,0} &= \mbs{\beta}_{0,k}^\text{gyr,0} + \Delta t \mbf{w}_{0,k}^\text{gyr,0}, \label{eq:gyr_0_process}\\
    \mbs{\beta}_{i,k+1}^\text{gyr,i} &= \mbs{\beta}_{i,k}^\text{gyr,i} + \Delta t \mbf{w}_{i,k}^\text{gyr,i}. \label{eq:gyr_i_process}
\end{align}

The relative accelerometer bias is more involved. The evolution of the individual accelerometer biases of the robots are also modelled as random walks, 
\begin{align}
    \mbs{\beta}_{0,k+1}^\text{acc,0} &= \mbs{\beta}_{0,k}^\text{acc,0} + \Delta t \mbf{w}_{0,k}^\text{acc,0}, \label{eq:acc_0_process}\\
    \mbs{\beta}_{i,k+1}^\text{acc,i} &= \mbs{\beta}_{i,k}^\text{acc,i} + \Delta t \mbf{w}_{i,k}^\text{acc,i}. \label{eq:acc_i_process}
\end{align}
The evolution of the relative accelerometer bias is a function of the individual accelerometer biases of the robots and the relative pose between the robots, and is given by
\begin{align}
    \mbs{\beta}_{0,k+1}^\text{acc,0i} &= \mbs{\beta}_{0,k+1}^\text{acc,0} - \mbs{\Pi} \mbf{T}_{0i,k+1} \mbs{\Pi}^\trans \mbs{\beta}_{i,k+1}^ \text{acc,i}, \label{eq:acc_0i_process}
\end{align}
where $\mbs{\Pi} \triangleq \bma{cc} \mbf{1}_{3} & \mbf{0}_{3 \times 2} \ema \in \mathbb{R}^{3 \times 5}$. Using \eqref{eq:process_model_dt_wBias}, \eqref{eq:acc_0_process}, and \eqref{eq:acc_i_process}, the relation in \eqref{eq:acc_0i_process} can be written as
\begin{align}
    \mbs{\beta}_{0,k+1}^\text{acc,0i} &= \textcolor{blue}{\mbs{\beta}_{0,k}^\text{acc,0} - \mbs{\Pi} \mbf{B}_{0,k}^{-1} \mbf{T}_{0i,k} \mbf{B}_{i,k} \mbs{\Pi}^\trans \mbs{\beta}_{i,k}^\text{acc,i}} + \Delta t \mbf{w}_{0,k}^\text{acc,0} - \mbs{\Pi} \mbf{B}_{0,k}^{-1} \mbf{T}_{0i,k} \mbf{B}_{i,k} \mbs{\Pi}^\trans \Delta t \mbf{w}_{i,k}^\text{acc,i} \label{eq:acc_0i_process_w_noise} \\
    &\approx \textcolor{blue}{\mbs{\beta}_{0,k}^\text{acc,0i}} + \Delta t \mbf{w}_{0,k}^\text{acc,0} - \mbs{\Pi} \mbf{B}_{0,k}^{-1} \mbf{T}_{0i,k} \mbf{B}_{i,k} \mbs{\Pi}^\trans \Delta t \mbf{w}_{i,k}^\text{acc,i}, \label{eq:acc_0i_process_approx}
\end{align}
where the lattermost approximation is dependent on an assumption that $\Delta t$ is sufficiently small. As $\Delta t \to 0$, it can be shown that $\mbf{B}_{0,k} \to \mbf{1}$ and $\mbf{B}_{i,k} \to \mbf{1}$, meaning that the first two components in \eqref{eq:acc_0i_process_w_noise} are approximately of the same form as the right-hand side in \eqref{eq:acc_0i_process} and can be combined into the relative acceleromer bias term $\mbs{\beta}_{0,k}^\text{acc,0i}$. Perturbing \eqref{eq:gyr_0_process}, \eqref{eq:gyr_i_process}, and \eqref{eq:acc_0i_process_approx} is then straightforward.

\section{Preintegration} \label{sec:preintegration}

The preintegration of the IMU measurements is also affected by the presence of IMU biases, but is quite similar to the preintegration shown in \cite[Section~VII]{Shalaby2023c}. In the proposed preintegration framework, each robot constructs its RMI by correcting the gyroscope measurements using its own gyroscope bias estimate and inflating the uncertainty associated with the RMI based on the uncertainty of the bias estimate. Nonetheless, each robot leaves the accelerometer measurements uncorrected when constructing the RMI.

To derive this, first note that
\begin{align}
    \label{eq:process_model_ell_to_m_bias}
    \mbf{T}_{0i,m} = \left(\prod_{k = \ell}^{m-1} \mbf{B}_{0,k}\right)^{-1} \mbf{T}_{0i, \ell} \prod_{k = \ell}^{m-1} \mbf{B}_{i,k},
\end{align}
meaning that the RMI constructed by Robot $i$ is of the form
\beq
    \Delta \mbf{T}_{i,\ell:m}^\text{B} = \prod_{k = \ell}^{m-1} \mbf{B}_{i,k} \in DE_2(3). \nonumber
\eeq
Therefore, \eqref{eq:process_model_ell_to_m_bias} can be written as 
\beq
    \mbf{T}_{0i,m} = \left(\prod_{k = \ell}^{m-1} \mbf{B}_{0,k}\right)^{-1} \mbf{T}_{0i, \ell} \Delta \mbf{T}_{i,\ell:m}^\text{B}, 
\eeq
which differs from the RMI in \cite[Section~VII]{Shalaby2023c} in that the gyroscope measurements are corrected using the Robot $i$'s estimate of its own gyroscope bias. Consequently, the RMI can be updated iteratively as 
\beq
    \Delta \mbf{T}_{i,\ell:k+1}^\text{B} = \Delta \mbf{T}_{i,\ell:k}^\text{B} \mbf{B}_{i,k}.
\eeq
As in \cite{Shalaby2023c}, a perturbation of the form
\beq
    \Delta \mbf{T}_{i,\ell:m}^\text{B} = \Delta \mbfbar{T}_{i,\ell:m}^\text{B} \operatorname{Exp} (\mbfdel{w}_{i,\ell:m}^\text{B}) \nonumber
\eeq
is defined for the RMI, and using \eqref{eq:Bi_perturbation}, the perturbation of the RMI is given by
\begin{align}
    \mbfdel{w}_{i,\ell:k+1}^\text{B} = \operatorname{Ad}(\bar{\mbf{B}}_{i,k}^{-1}) \mbfdel{w}_{i,\ell: k}^\text{B} + \mbf{L}_{i,k}^\text{B} \delta \mbf{u}_{i,k} - \mbf{L}_{i,k}^\text{B} \mbf{E} \delta \mbs{\beta}_{i,k}^{\text{gyr},i},
\end{align}
where the last term reflects the increased uncertainty of the RMI associated with the uncertainty in the gyroscope bias estimate.

Lastly, the asynchronous-input filter shown in \cite[Section~VII.C]{Shalaby2023c} can now be formulated for bias-modelling applications. At time-steps where there is no communication with the neighbour, the pose and bias process models are given by
\begin{align}    
    \mbc{T}_{0i,k+1} &= \mbf{B}_{0,k}^{-1} \mbf{T}_{0i,k}, \qquad \mbc{T}_{0i,k+1} \in DE_2(3), \\
    \mbs{\beta}_{0,k+1}^\text{gyr,0} &= \mbs{\beta}_{0,k}^\text{gyr,0} + \Delta t \mbf{w}_{0,k}^\text{gyr,0}, \\
    \mbs{\beta}_{0,k+1}^\text{acc,0i} &= \mbs{\beta}_{0,k}^\text{acc,0i} + \Delta t \mbf{w}_{0,k}^\text{acc,0} - \mbs{\Pi} \mbf{B}_{0,k}^{-1} \mbf{T}_{0i,k} \mbs{\Pi}^\trans \Delta t \mbf{w}_{i,k}^\text{acc,i}.
\end{align}
Meanwhile, at time-steps when Robot $i$ sends the RMI $\Delta \mbf{T}_{i,\ell:m}^\text{B}$, the pose and bias process models are given by
\begin{align}    
    \mbf{T}_{0i,m} &= \mbf{B}_{0,m-1}^{-1} \mbc{T}_{0i,m-1} \Delta \mbf{T}_{i,\ell:m}^\text{B}, \\
    \mbs{\beta}_{0,m}^\text{gyr,0} &= \mbs{\beta}_{0,m-1}^\text{gyr,0} + \Delta t \mbf{w}_{0,m-1}^\text{gyr,0}, \\
    \mbs{\beta}_{0,m}^\text{acc,0i} &= \mbs{\beta}_{0,m-1}^\text{acc,0i} + \Delta t \mbf{w}_{0,m-1}^\text{acc,0} - \mbs{\Pi} \mbf{B}_{0,m-1}^{-1} \mbc{T}_{0i,m-1} \Delta \mbf{T}_{i,\ell:m}^\text{B} \mbs{\Pi}^\trans \Delta t \mbf{w}_{i,m-1}^\text{acc,i}
\end{align}
These equations can then be perturbed in a manner similar to \cite{Shalaby2023c}.

\section{Simulation Results} \label{sec:simulation}

\begin{table}[H]
    \renewcommand{\arraystretch}{1.2}
    \caption{\vspace{-5pt}Bias simulation parameters. Other simulation parameters remain unchanged from \cite[Table I]{Shalaby2023c}.}
    \label{tab:sim_params_bias}
    \centering
    \begin{tabular}{c|c}
    \bfseries Specification & \bfseries Value\\
    \hline
    Accelerometer bias random walk std. dev. [m/s$^2$] & $1.58 \times 10^{-3}$  \\
    Gyroscope bias random walk std. dev. [rad/s] & $2.5 \times 10^{-5}$
    \end{tabular}
\end{table}

In order to validate the proposed framework in the presence of IMU biases, the same simulation runs as in \cite{Shalaby2023c} are repeated, but with the addition of IMU biases. The bias simulation parameters are given in Table \ref{tab:sim_params_bias}. Given that neighbours use their own gyroscope bias estimates to correct their gyroscope measurement before constructing the RMI, additional noise is added to the gyroscope bias true state of neighbours to simulate uncertain gyroscope bias estimates. This is then used to correct the gyroscope measurements and to inflate the RMI. 

\begin{figure*}[h]
    \centering
    \begin{minipage}{0.485\textwidth}%
        \centering
        \includegraphics[width=\textwidth]{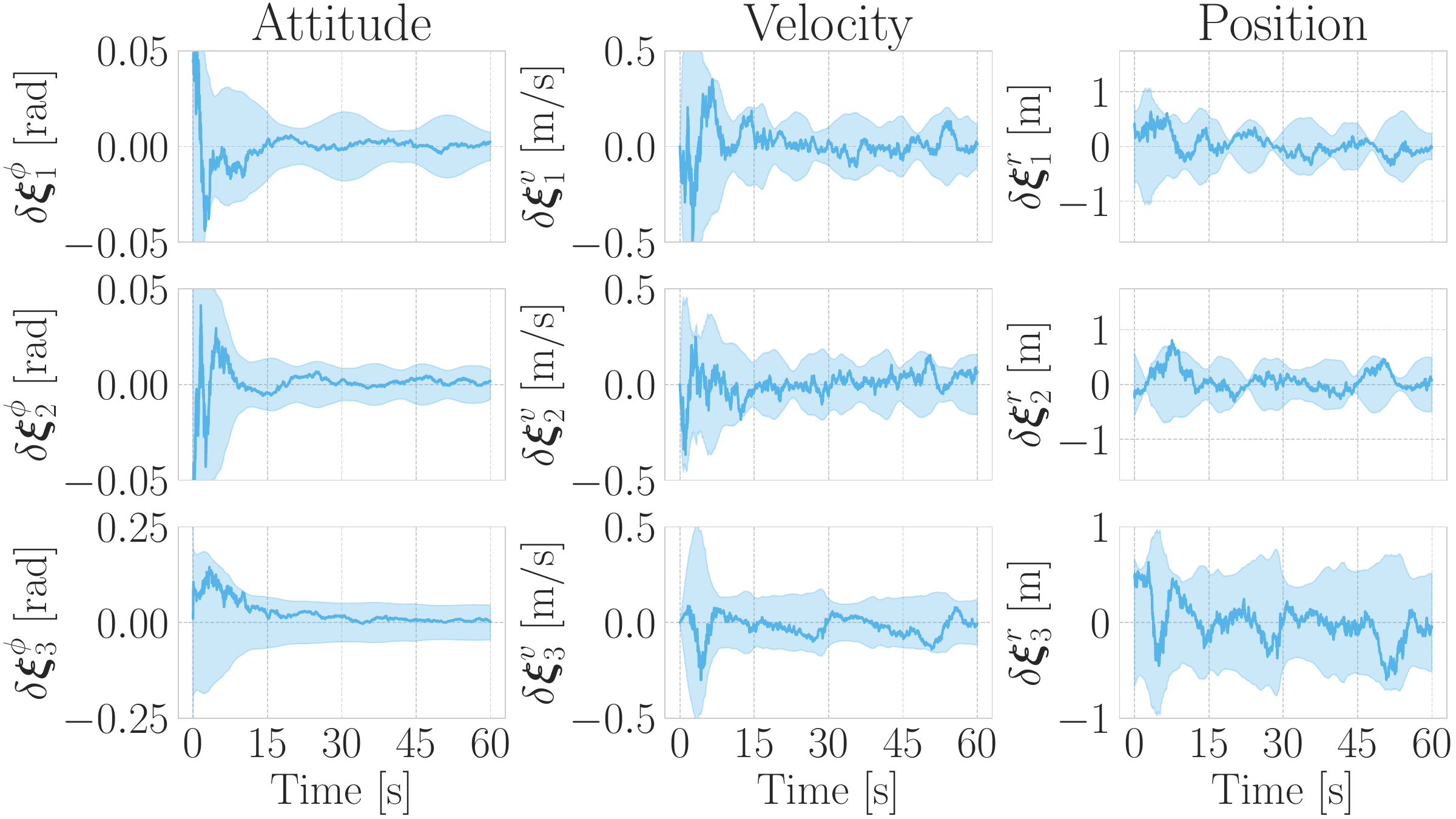}
        \label{fig:sim_pose_3sigma_bias}
    \end{minipage}\quad%
    \begin{minipage}{0.485\textwidth}%
        \centering
        \includegraphics[width=\textwidth]{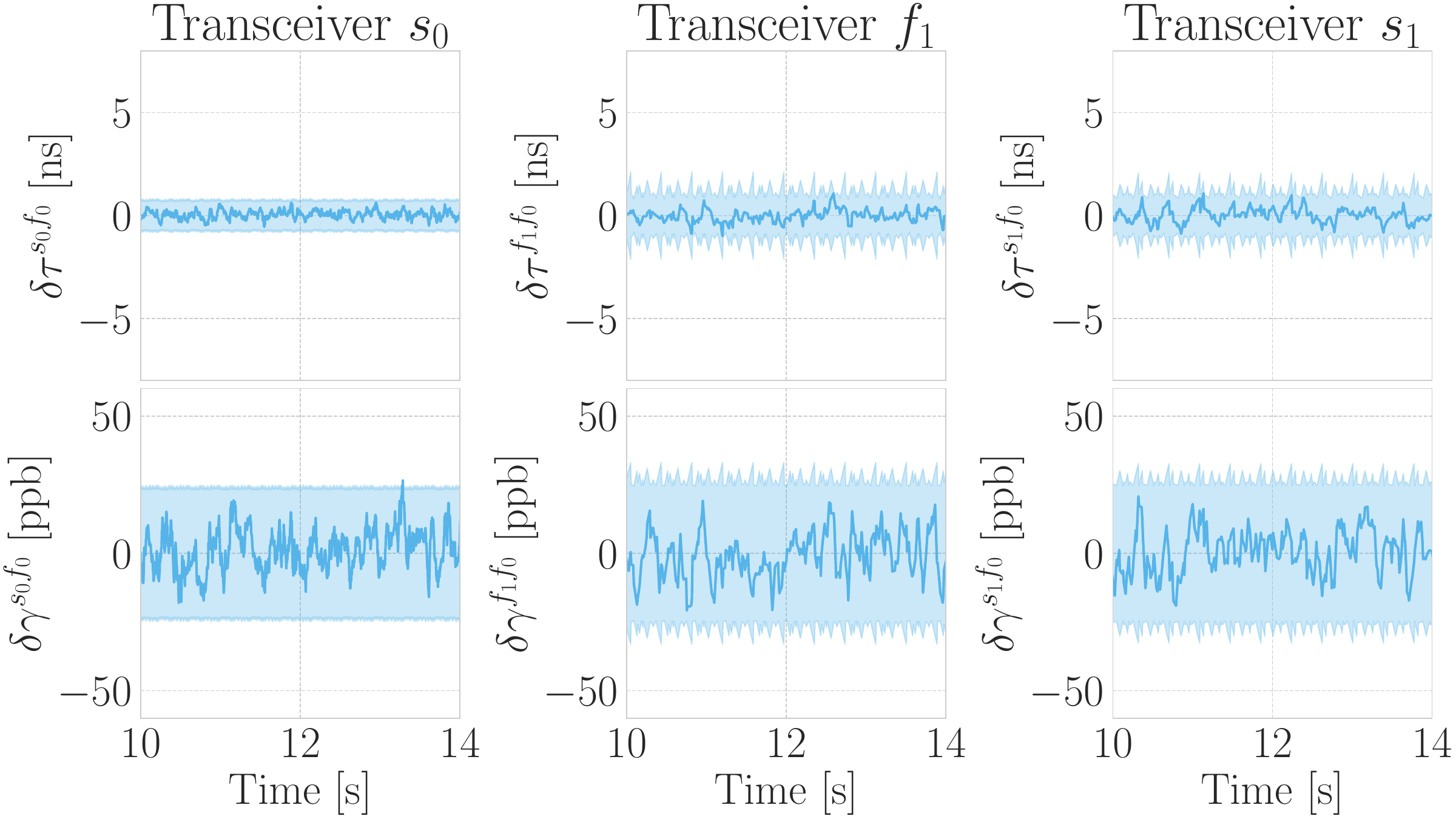}
        \label{fig:sim_clock_3sigma_bias}
    \end{minipage}
    \begin{minipage}{0.485\textwidth}%
        \centering
        \includegraphics[width=\textwidth]{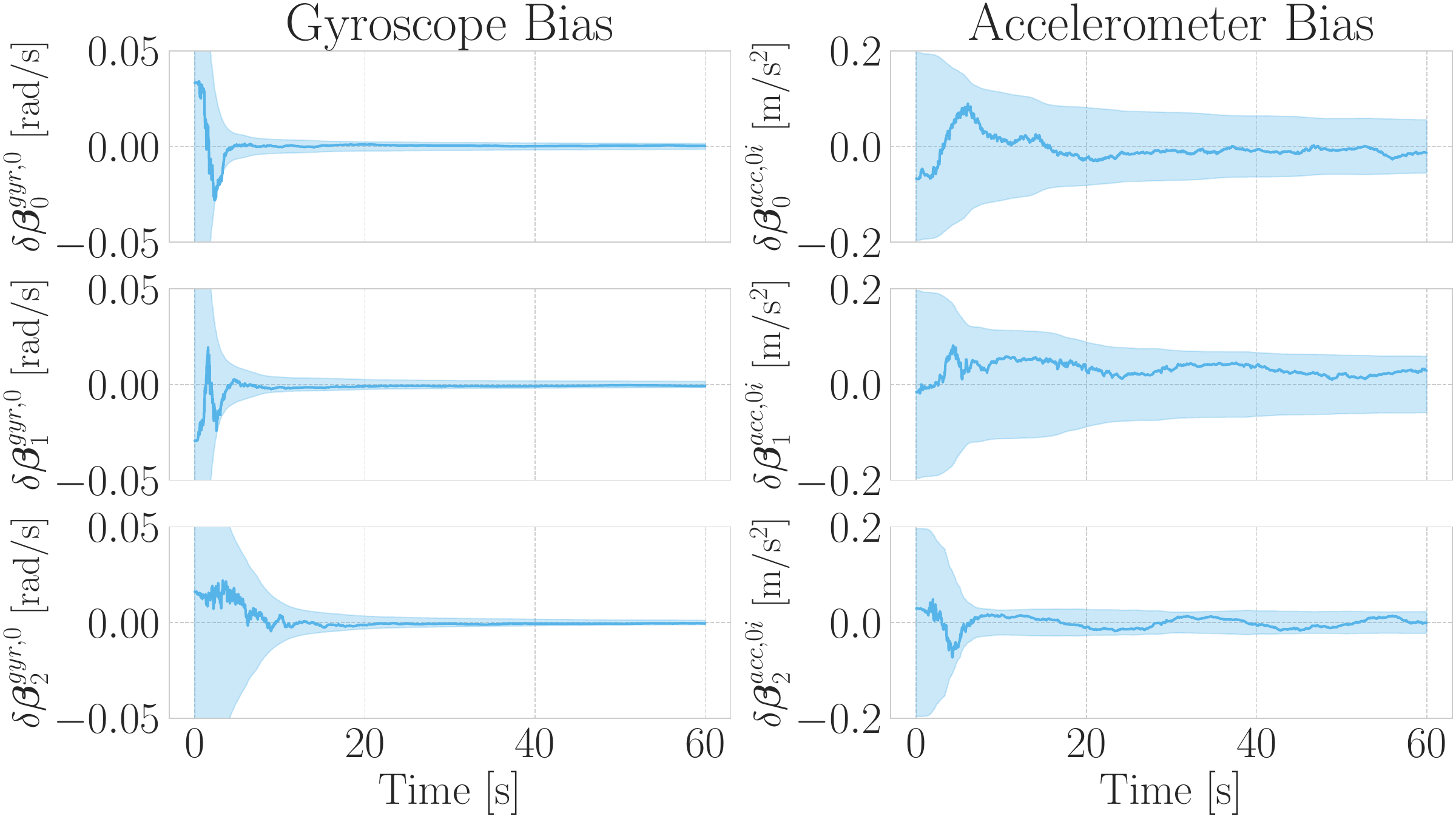}
        \vspace{-13pt}
        \label{fig:sim_bias_3sigma_bias}
    \end{minipage}
    \caption{Error plots and $\pm3\sigma$ bounds (shaded region) for Robot 0's estimate of Robot 1's relative pose, its own gyroscope bias, and Robot 1's relative accelerometer bias for Simulation S1.}
    \label{fig:sim_pose_bias}
\end{figure*}

\begin{figure}[h!]
    \centering
    \includegraphics[width=0.75\columnwidth]{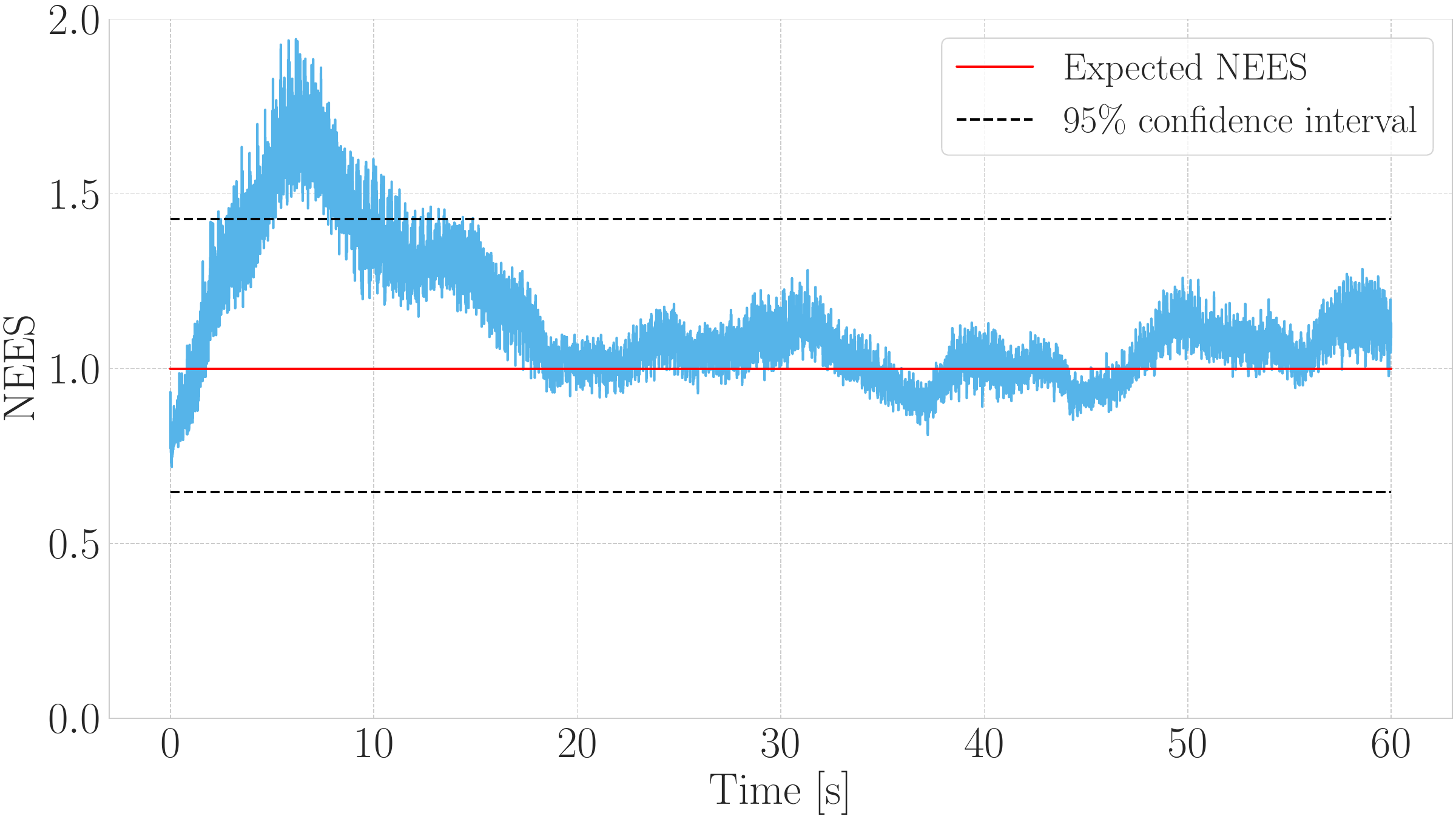}
    \caption{50-trial NEES plot for the proposed estimator on Simulation S3 in the presence of bias.}
    \label{fig:nees_bias}
\end{figure}

The results for Simulation S1 are shown in Figure~\ref{fig:sim_pose_bias}, where it can be seen that the gyroscope and relative accelerometer biases estimated by the reference robot do converge to the true values. Additionally, Simulation S3 is run to assess the consistency of the proposed estimator in the presence of IMU biases. The NEES plot for this simulation is shown in Figure~\ref{fig:nees_bias}, where it can be seen that the NEES values display a similar behaviour to \cite[Figure 12]{Shalaby2023c}, starting with weak observability and then converging towards consistency.

\section{Experimental Results} \label{sec:experimental}

\begin{figure*}[t!]
    \centering
    \begin{minipage}{0.49\textwidth}%
        \centering
        \subfloat[Subfigure 1 list of figures text][Without bias estimation.]{
        \includegraphics[width=\textwidth]{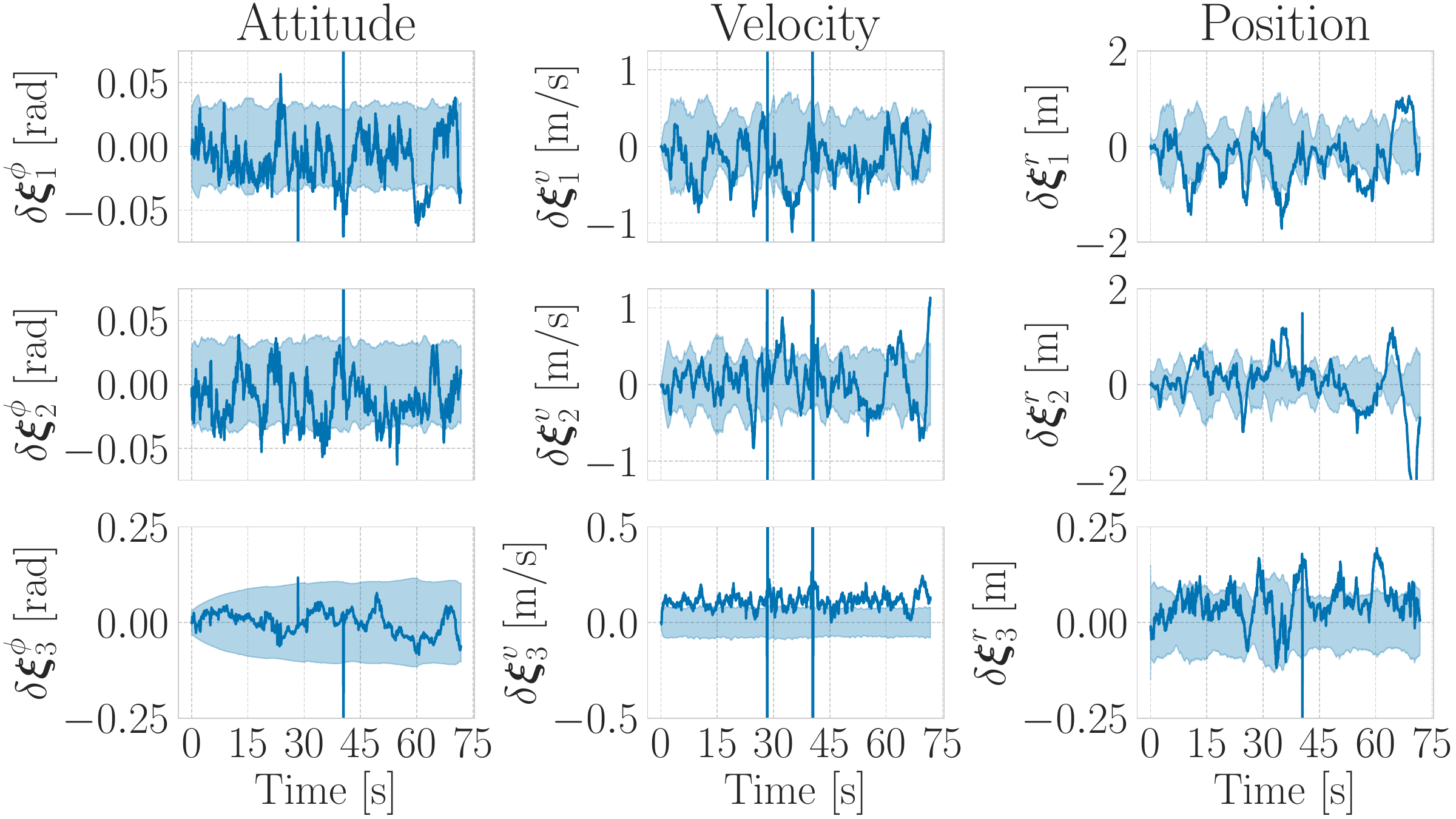}
        \label{fig:exp_pose_3sigma_noBias}}
    \end{minipage}
    \begin{minipage}{0.49\textwidth}%
        \centering
        \subfloat[Subfigure 2 list of figures text][With bias estimation.]{
        \includegraphics[width=\textwidth]{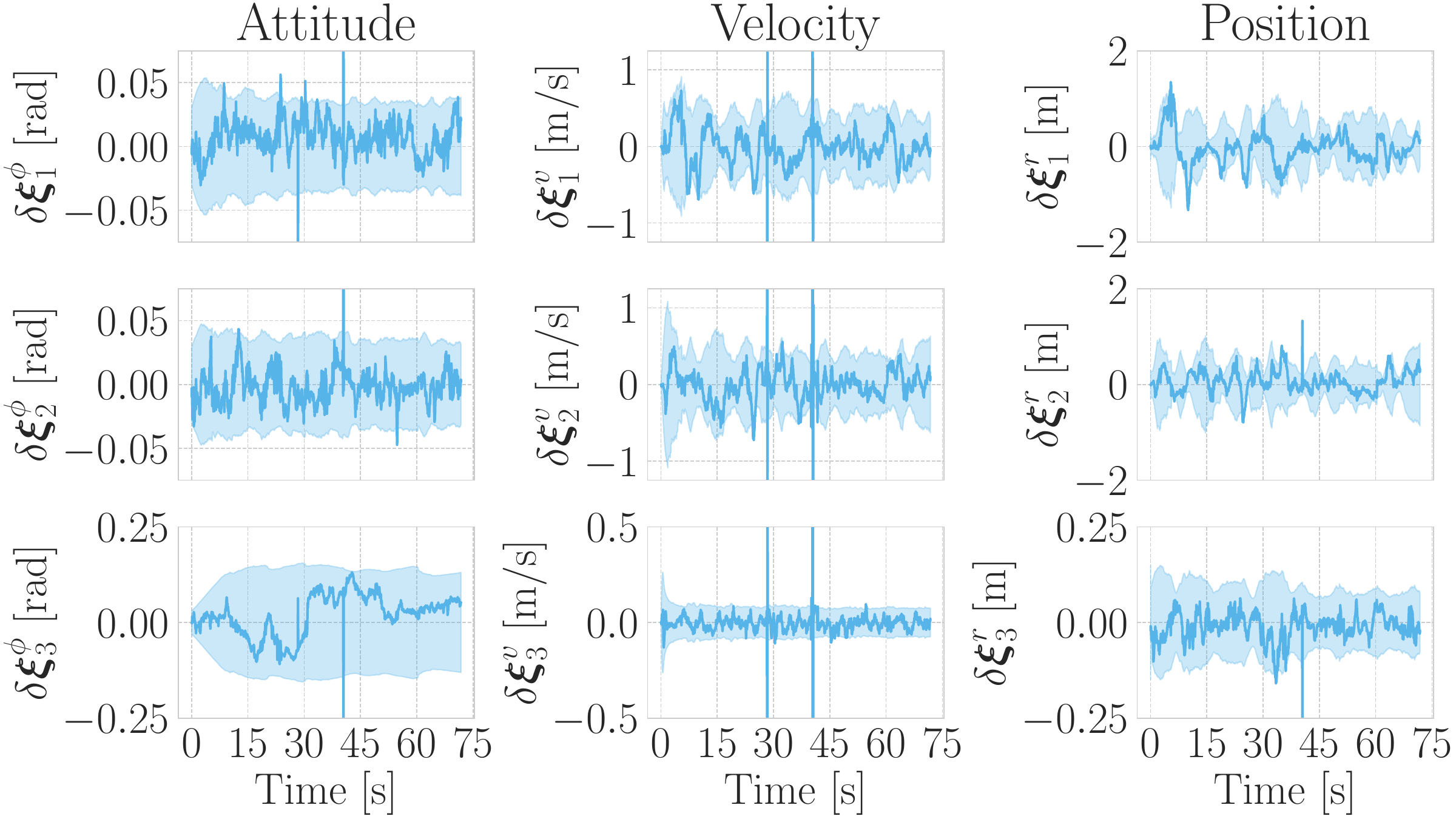}
        \label{fig:exp_pose_3sigma_wBias}}
    \end{minipage}
    \caption{Error plots and $\pm3\sigma$ bounds (shaded region) for Robot 0's estimate of Robot 1's relative pose for experimental trial 1, without offline bias correction.}
    \label{fig:exp_pose_bias}
\end{figure*}

The experimental results are also rerun with IMU bias estimation to validate the approach proposed in this document. These results differ from the results presented in \cite{Shalaby2023c} in that the IMU biases are no longer initialized using the motion capture system, except for the gyroscope biases of neighbouring robots to simulate neighbours running their own estimator. The proposed framework presented in \cite{Shalaby2023c} is then compared to the one presented here with IMU bias estimation, and it is shown that estimating biases in the absence of bias initialization does indeed improve performance, as shown in Figure \ref{fig:exp_pose_bias} for Trial 1 and in Table \ref{tab:exp_rmse_bias} for all trials. Note that the bias error plots are not shown as the true IMU bias is unknown. The performance with bias estimation is also comparable to the performance of the estimator in \cite{Shalaby2023c} with bias initialization, but is in fact typically worse probably due to the transient of the bias estimates before convergence, thus resulting in more uncertain pose estimates during the earlier stages. It is expected that for longer trajectories the performance of the estimator with bias estimation will be better than the estimator in \cite{Shalaby2023c} with bias initialization, as the initial bias estimates becomes less accurate with the progress of time. 

\newcolumntype{C}{>{\centering\arraybackslash}m{1.35cm}}
\newcolumntype{L}{>{\centering\arraybackslash}m{2cm}}
\begin{table*}
    \footnotesize
    \renewcommand{\arraystretch}{1.2}
    \caption{The RMSE of Robot 0's estimate of neighbouring robots' relative pose for multiple experimental trials, without offline bias correction.}
    \label{tab:exp_rmse_bias}
    \centering
    \begin{tabular}{C|LL|LL}
    & \multicolumn{2}{c}{\bfseries Robot 1} & \multicolumn{2}{c}{\bfseries Robot 2} \\
    \hline 
    & \bfseries Without Bias Estimation RMSE [m] & \bfseries With Bias Estimation RMSE [m] 
    & \bfseries Without Bias Estimation RMSE [m] & \bfseries With Bias Estimation RMSE [m] \\
    \hline
    Trial 1 & 0.785 & \bfseries 0.404 & 0.723 & \bfseries 0.410  \\
    Trial 2 & 1.232 & \bfseries 0.828 & 0.902 & \bfseries 0.615  \\
    Trial 3 & 0.916 & \bfseries 0.548 & 0.649 & \bfseries 0.413  \\
    Trial 4 & 1.282 & \bfseries 0.753 & 0.853 & \bfseries 0.638  \\
    \end{tabular}
    \vspace{-10pt}
\end{table*}

\section{Conclusion}

In this document, the proposed estimator in \cite{Shalaby2023c} is extended to include IMU bias estimation. It is shown that, when modelling the evolution of biases as a random walk, the IMU biases can be incorporated into the process model while still maintaining the Sylvester equation form. To do so, each robot estimates its own gyroscope bias in its own body frame, and uses this estimate to correct the IMU measurements and inflating the covariance when constructing the RMI. Additionally, each robot estimates a relative accelerometer bias to every neighbour in the robot's own body frame, which does not affect the computed RMI. The proposed estimator is then validated in simulation and in experiments, and is shown to improve performance in the absence of bias initialization.

\clearpage

\addcontentsline{toc}{section}{References}
\bibliographystyle{ieeetr}
\bibliography{ref}
\end{document}